%% file: cnsm2020.tex
\documentclass[10pt, conference]{IEEEtran}

\usepackage[latin1]{inputenc}  
\usepackage{graphicx,url}
\usepackage{multirow}  
\usepackage{hhline}
\usepackage{graphicx}
\usepackage{subfigure,amssymb,amsmath}
\usepackage{color,soul}
\usepackage[dvipsnames]{xcolor}
\usepackage{bbm}
\usepackage{float}
\makeatletter  
\newif\if@restonecol  
\makeatother

\usepackage[linesnumbered,ruled,vlined]{algorithm2e}
\usepackage{algpseudocode}  
\usepackage{amsmath}  
\usepackage{fancyhdr} 
\usepackage{tabularx}
\usepackage{comment} 

\renewcommand\thetable{\arabic{table}}
\IEEEoverridecommandlockouts
\IEEEpubid{\makebox[\columnwidth]{978-3-903176-31-7~\copyright~2020 IFIP\hfill} \hspace{\columnsep}\makebox[\columnwidth]{ }}

\usepackage{etoolbox}
\makeatletter
\patchcmd{\@makecaption}
  {\scshape}
  {}
  {}
  {}
\makeatother

\begin{document}


\title{Online feature selection for rapid, low-overhead learning in networked systems}

\author{\IEEEauthorblockN{Xiaoxuan Wang \IEEEauthorrefmark{2}  Forough Shahab Samani \IEEEauthorrefmark{2}\IEEEauthorrefmark{3} and 
 Rolf Stadler\IEEEauthorrefmark{2}\IEEEauthorrefmark{3}}

 \IEEEauthorblockA{\IEEEauthorrefmark{2}
Dept. of Computer Science, KTH Royal Institute of Technology, Sweden
 }
 \IEEEauthorblockA{\IEEEauthorrefmark{3} RISE Research Institutes of Sweden \\
  \newline
Email: \{xiaoxuan, foro, stadler\}@kth.se
\\
\today
}
}

\maketitle

\thispagestyle{plain}
\pagestyle{plain}

\input{abstract}

\input{keywords} 

\input{introduction}

\input{problem_formulation}

\input{Creating_ranked_feature_lists}

\input{Testbed_and_trace_for_evaluation}

\input{Computin_stabl_feature_sets}

\input{online_feature_selection_with_low_overhead}

\input{related_work}

\input{conclusions_and_future_work}

\input{acknowledgments}

\bibliographystyle{IEEEtran}
\bibliography{cnsm2020}

\end{document}

%% file: abstract.tex
\begin{abstract}
\label{sec:abstract}
Data-driven functions for operation and management often require measurements collected through monitoring for model training and prediction. The number of data sources can be very large, which requires a significant communication and computing overhead to continuously extract and collect this data, as well as to train and update the machine-learning models.  We  present  an  online  algorithm,  called OSFS,  that selects  a  small feature set from a large number of available data sources,  which  allows for  rapid,  low-overhead,  and  effective  learning and prediction.  OSFS is instantiated with a feature ranking algorithm and applies the  concept of a stable feature set, which we introduce in the paper. We perform extensive, experimental evaluation of our method on data from an in-house testbed. We find that OSFS requires several hundreds  measurements to reduce the number of data sources by two orders of magnitude, from which models are trained with acceptable prediction accuracy.  
While our method is heuristic and can be improved in many ways, the results clearly suggests that many learning tasks do not require a lengthy monitoring  phase and expensive offline training.

\end{abstract}

%% file: keywords.tex

%% file: introduction.tex
\section{Introduction}
\label{sec:introduction}
Data-driven network and systems engineering is based upon applying AI/ML methods to data collected from an infrastructure in order to build novel functionality and management capabilities. This is achieved through learning tasks that use this data for training. Examples are KPI prediction and forecasting through regression and anomaly detection through clustering techniques.

Data sources that feed the learning tasks include system logs, telemetry data, and real-time measurements collected through monitoring. The number of available data sources can be very high, even in small systems. For example, on our testbed at KTH, which includes 10 compute servers, we can extract several thousand metrics from the operating system and orchestration layers. Since these metrics are dynamic, we monitor them periodically, e.g., once per second. It requires a significant overhead to extract and collect this data, as well as a significant computational overhead to train and update the machine-learning models that underlie the learning tasks. Considering the fact that the monitoring and computational overhead increases at least linearly with the number of measurements and the dimensionality of the input, i.e., the number of (one-dimensional) data sources, it becomes vital to reduce the number of data sources to the extent possible.

The focus of this paper is on a novel online source-selection method that requires only a small number of measurements to significantly reduce the number of sources needed for training models that are effective for learning tasks. As a result, the communication overhead for monitoring and the computational overhead and time needed for the model training are significantly reduced. 

Using the terminology of machine learning, we call a (one-dimensional, scalar) data source also \emph{a feature}, and we refer to measurements taken from a set of data sources at a specific time as \emph{a sample}.

Our approach consists of (1) ranking the available data sources using (unsupervised) feature selection algorithms and (2) identifying \emph{stable feature} sets that include only the top $k$ features.  We call a feature set stable, if it remains sufficiently similar when additional samples are considered. 

We evaluate our approach using traces from an in-house testbed that runs two services under different load conditions. The results show that our method can reduce the number of data sources needed for learning tasks by two orders of magnitude, while still achieving acceptable errors for a prediction task. The reduction in input dimensionality is consistent with results from our earlier work, which studies non-linear methods for dimensionality reduction in an offline setting \cite{samani2019efficient}. We find that a stable feature set can often be identified with only a few hundred samples. As a consequence, the monitoring effort required for collecting the data for a machine-learning task can be reduced quickly. We consider this capability key for increasing the acceptance of data-driven engineering solutions.  

With this paper, we make the following contributions:
\begin{itemize}
    \item We present an online algorithm, which we call OSFS, that selects a small feature set from a large number of available data sources using a small number of measurements, which allows for rapid, low-overhead, and effective learning. The algorithm is initialized with a feature ranking algorithm and applies the concept of a stable feature set, which we introduce in the paper.
    \item We perform an extensive, experimental evaluation of our method on an in-house testbed.  
\end{itemize}

The significance of our findings lies in the prospect that many data-driven functions in networked systems can be trained rapidly and with low overhead and thus do not require a lengthy monitoring phase and expensive offline training. 

This paper contains results from a master thesis project conducted at KTH \cite{xiaoxuans_thesis}.

The rest of the paper is organized as follows. Section \ref{sec:problem_formulation} formulates the problem we address in the paper. Section  \ref{sec:creating_ranked_feature_lists}
describes the feature selection methods we use to obtain ranked feature lists. Section \ref{sec:testbed} details our testbed, the experiments we conduct, the measurements we collect during experiments, and the traces we generate from this data. Section \ref{sec:Computing_stable_feature_sets} introduces the concept of the stable feature set. Section \ref{sec:online_feature_selection_with_low_overhead} presents our online features selection method and evaluates the method using testbed traces. Section \ref{sec:related_work} surveys related work. Finally, Section \ref{sec:conclusions} presents the conclusions and future work.

%% file: problem_formulation.tex
\section{Problem formulation and approach}
\label{sec:problem_formulation}

We consider a monitoring infrastructure that collects readings from a set $\emph{\textbf{F}}$ of $n$ distributed data sources (or features). Each feature has a one-dimensional, numerical value that changes over time. We collect readings at discrete times $t$ and store them in sample vectors $\emph{\textbf{X}}_t \in \mathbb{R}^n,  t=1,2,3,..$ . Our plan is to identify a subset $\emph{\textbf{F}}_{k} \subset \emph{\textbf{F}}$ with $k \ll n$ features using the samples $\emph{\textbf{X}}_1, \emph{\textbf{X}}_2, ..,  \emph{\textbf{X}}_{t_k}$.

Second, we consider a learning task, like KPI prediction or anomaly detection, whose model is trained using the samples $\emph{\textbf{X}}_t \in \mathbb{R}^k$ with the features from  $\emph{\textbf{F}}_{k}$.
 
In order to keep low the monitoring overhead for collecting the samples and the computational overhead for training the model associated with the learning task, the numbers for $k$ and $t_k$ should be small. Note that $k$ indicates the number of data sources that need to be monitored to train the model and that $t_k$ refers to the number of measurements that are needed to compute $\emph{\textbf{F}}_{k}$. Assuming periodic measurements, $t_k$ further indicates the time it takes until the feature set $\emph{\textbf{F}}_{k}$ is available.

Our objective thus is to select $k$ and $t_k$ as small as possible, while enabling the models trained using $\emph{\textbf{F}}_{k}$ to be equally (or similarly) effective for prediction as those trained using the complete feature set $\emph{\textbf{F}}$ and a large number of samples.

The task of selecting a subset of features from a larger set is called \emph{feature selection}  in machine learning and data mining and is a well-studied topic area.
(See Section \ref{sec:related_work}). 
We are specifically interested in unsupervised feature selection methods, whereby the values of the target are not known during the feature selection process, i.e., the process to compute $\emph{\textbf{F}}_{k}$. This allows us to keep the feature selection process independent from the the learning task and will enable different learning tasks in a system to share the same feature subset.

The problem we address in this paper is to find an online algorithm that reads a sequence of $n$-dimensional sample vectors $\emph{\textbf{X}}_1, \emph{\textbf{X}}_2, .. $ one by one, computes $k$  and the feature set $\emph{\textbf{F}}_{k}$, and terminates after step $t_k$. The values for $k$ and $t_k$ should be small, while $\emph{\textbf{F}}_{k}$ must be effective in training models for learning tasks.

In our approach, we choose an unsupervised feature selection method that ranks the n features in every step $t$ of the online algorithm and checks how the top $i$ features ($i=1,..,n$) change with increasing $t$. We introduce a similarity metric that captures this change (see Section \ref{sec:Computing_stable_feature_sets}). If the similarity between the top $k$ features in consecutive steps is high and not increasing anymore, the algorithm terminates and the values for $\emph{\textbf{F}}_{k}$, $k$ and $t_k$ are returned (see Section \ref{sec:online_feature_selection_with_low_overhead}). 

Note that we assume here that the feature set $\emph{\textbf{F}}$ is fixed. In a real system that runs over some time, changes to the physical configuration or the virtualization layer occur, which result in changes to the set of available measurement points, i.e. the feature set. In such a case, the online algorithm must be re-started. Note also that we do not investigate in this work how many samples are needed to train the model of the learning task. This will be done in future work.

%% file: creating_ranked_feature_lists.tex
\section{Creating ranked feature lists}
\label{sec:creating_ranked_feature_lists}

In this section, we describe two algorithms (ARR, LS) that produce a ranked feature list from a list of samples. They are based on unsupervised feature selection methods from the literature. In addition, we include a third ranking algorithm (TB), which is supervised and will serve as a baseline.  We will evaluate the suitability of theses algorithms for our online feature selection method OSFS in Section \ref{sec:online_feature_selection_with_low_overhead}. 

Table \ref{notation} shows the notation we use in the paper. The available data for computing the feature set $\emph{\textbf{F}}_{k}$ is presented as a design matrix $\emph{\textbf{X}} \in \mathbb{R}^{m\times n}$, whose $n$ columns represent the feature vectors and $m$ rows represent the samples in the data set. Since we assume that the samples arrive in sequence one-by-one, $m$ is increasing over time and can be interpreted as time index. 

\begin{table}[ht]
    \centering
    \caption{Table of notation}
    \label{notation}
    \scalebox{0.9}{
    \begin{tabular}{|c|c|}
    \hline
    $\emph{\textbf{X}}$& data set \\
    \hline 
    $n$& number of features \\
    \hline 
    $m$& number of samples \\
    \hline 
     $\emph{\textbf{X}}_{i,:}\ $or $\emph{\textbf{X}}_{i}    (i=1,...,m)$ & $i$-th row or $i$-th sample of $\emph{\textbf{X}}$\\
    \hline
    $\emph{\textbf{X}}_{:,j}\ (j=1,...,n)$& $j$-th column or $j$-th feature vector of $\emph{\textbf{X}}$\\
    \hline
    $\mathit{X_{i,j}}$& element of the $i$-th row and the $j$-th column of $\emph{\textbf{X}}$\\
    \hline
    $k$& number of selected features \\
    \hline 
    $t_k$& number of samples used for feature selection\\
    \hline
    $\emph{\textbf{F}}$& set of all available features\\
    \hline 
    $\emph{\textbf{F}}_{k}$& subset with k selected features\\ 
    \hline
    \end{tabular}}
\end{table}

The first algorithm is Adapted Relevance Redundancy Feature Selection (ARR). It is based on the Relevance Redundancy Feature Selection (RRFS) method \cite{ferreira2012}, which we adapted to compute a ranked feature list. ARR uses two criteria in assessing the rank of a feature: (a) relevance, which relates to the distance of a feature vector to the mean of all feature vector vectors, and (b) redundancy, which relates to the cosine similarity between a feature vector and the vectors of all other features. High relevance and low similarity result in a high score. 
The pseudo-code of ARR is given in Algorithm \ref{arr_algorithm}.  First, the relevance of a feature is computed as the mean absolute difference of its feature vector from the mean (line 3). Then, the relevance of a feature is computed as the sum of the cosine similarity of its feature vector and each feature vector of the data set (lines 5-6). The score for ranking a feature is the relevance value divided by the redundancy value (line 7). The computational complexity of ARR is $O(n^2m)$.
Recall that the cosine similarity between two features vectors $\emph{\textbf{X}}_{:,a}$ and $\emph{\textbf{X}}_{:,b}$ is calculated as:
\begin{equation}
    cosim(\emph{\textbf{X}}_{:,a},\emph{\textbf{X}}_{:,b})=\left | \frac{\sum_{i=1}^{m}(\mathit{X_{i,a}}\mathit{X_{i,b}})}{(\sqrt{\sum_{i=1}^{m}\mathit{X_{i,a}}^{2}})(\sqrt{\sum_{i=1}^{m}\mathit{X_{i,b}}^{2}})}\right |
\end{equation}

\begin{algorithm} 
  \label{arr_algorithm}
  \caption{Adapted Relevance Redundancy Feature Selection (ARR)}  
  \KwIn{Data matrix $\emph{\textbf{X}} \in \mathbb{R}^{m\times n}$}  
  \KwOut{Ranked feature list $\emph{\textbf{F}}^{'}$}
  $\emph{\textbf{F}}^{'}$=[]\;
  \For{$i=1;i\leq n;i++$} 
  {  
    $relevance_{i}=\sum_{j=1}^{m}\left | \mathit{X_{j,i}}-\overline{\emph{\textbf{X}}_{:,i} }\right |$\;  
    $sim\_sum_{i}=0$\;  
    \For{$j=1;j\leq n;j++$}  
    {  
      $sim\_sum_{i}+=cosim(\emph{\textbf{X}}_{:,i},\emph{\textbf{X}}_{:,j})$\;  
    }  
    $score_{i}=\frac{relevance_{i}}{sim\_sum_{i}}$\;
  } 
   Construct $\emph{\textbf{F}}^{'}$ as list of all features sorted by $score_{i}$ in descending order\;
  return $\emph{\textbf{F}}^{'}$\;  
\end{algorithm}  

The second algorithm is Laplacian Score (LS) \cite{he2006laplacian}. It follows the so-called filter method which examines intrinsic properties of the data to evaluate the features. LS ranks those features high that preserve locality with respect to a neighborhood graph. Algorithm 2 shows the pseudocode of LS. Input parameters for this algorithm are the design matrix and the number of local neighbors $K$.
First, using the $m$ sample vectors as the nodes of the graph, a neighborhood graph is constructed with the $K$ nearest neighbors of each node as the links of the graph (line 1). Then, the graph connectivity and the distance between node pairs is used to compute the weight matrix (line 2). The graph Laplacian matrix is computed in line 4. The Laplacian score of all $n$ features is obtained in the for loop (lines 5-7). A low score for a feature signifies high locality preservation (see \cite{he2006laplacian} for justification), and the features are ranked according to increasing score (line 8). The computational complexity of the LS algorithm is $O(nm^2)$.

In the evaluations reported in Sections V and VI of this paper, we choose $K$ in relation to the value of $m$: $K=2$ for $0<m\leq 16$; $K=5$ for $16<m\leq 128$; $K=10$ for $m>128.$

\begin{algorithm} 
  \label{ls_algorithm}
  \caption{Laplacian Score (LS)}  
  \KwIn{Data matrix $\emph{\textbf{X}} \in \mathbb{R}^{m\times n}$, $K \in \mathbb{N}$ nearest neighbors}  
  \KwOut{Ranked feature list $\emph{\textbf{F}}^{'}$}
  
  Construct graph $\emph{\textbf{G}}$ of $K$ nearest neighbors from nodes $\emph{\textbf{X}}_{i,:}, i=1,..,m$\;
  Compute weight matrix $\emph{\textbf{S}} \in \mathbb{R}^{m\times m}$ from $\emph{\textbf{G}}$
  \begin{equation*}
  S_{ij}=\left\{\begin{matrix}
   e^{-\left \| \emph{\textbf{X}}_{i,:}-\emph{\textbf{X}}_{j,:} \right \|^{2}} & if\ nodes \ i \ and \ j \ are\ connected,\\ 
   0& otherwise
    \end{matrix}\right.
    \end{equation*}\\
  $\emph{\textbf{D}}= diag(\emph{\textbf{S1}})$ where 
  $\emph{\textbf{1}}=\left [ 1,...,1 \right ]^{T}$,
  $\emph{\textbf{D}}\in \mathbb{R}^{m\times m}$\;
  $\emph{\textbf{L}}= \emph{\textbf{D}} - \emph{\textbf{S}}$\;
  
  \For{$i=1;i\leq n;i++$} 
  {  
    $\emph{\textbf{V}}_{i}=\emph{\textbf{X}}_{:,i}-\frac{\emph{\textbf{X}}_{:,i}^{T}\emph{\textbf{D}}\emph{\textbf{1}}}{\emph{\textbf{1}}^{T}\emph{\textbf{D}}\emph{\textbf{1}}}\emph{\textbf{1}}$ where
    $\emph{\textbf{V}}_{i}\in \mathbb{R}^{m\times 1}$\;  
    $lscore_{i}=\frac{\emph{\textbf{V}}_{i}^{T}\emph{\textbf{L}}\emph{\textbf{V}}_{i}}{\emph{\textbf{V}}_{i}^{T}\emph{\textbf{D}}\emph{\textbf{V}}_{i}}$\;
  } 
   Construct $\emph{\textbf{F}}^{'}$ as list of all features sorted by $lscore_{i}$ in ascending order\;
  return $\emph{\textbf{F}}^{'}$\;  
\end{algorithm} 

The third algorithm for feature ranking is tree-based feature selection (TB). TB is a supervised algorithm in which the feature importance is computed using random forest prediction \cite{hastie2009elements} \cite{deng2012feature}. We consider TB a baseline to better assess the performance of ARR and LS. In this work, scikit-learn library is used to compute the TB feature ranking \cite{scikitFeatureSelection} \cite{scikitFeatureImportance}. We set the number of trees to 100 and use default values for the remaining parameters. The computational complexity of TB is $O(Tnm\log _{2}m)$ where $T$ is the number of trees, $n$ is the number of features, and $m$ is the number of samples \cite{TBcomplexity}. 

%% file: testbed_and_trace_for_evaluation.tex
\section{Testbed and traces for evaluation}
\label{sec:testbed}

\subsection{Testbed and services}
\label{subsec:testbed_and_services}
In this section, we describe the experimental infrastructure and the structure of the data traces that we create. Further, we describe the services that run on this infrastructure, namely, a Video-on-Demand (VoD) service and Key-Value (KV) store. Lastly, we explain the load patterns we use and the experiments we run to obtain the traces. 

\label{sec:testbed-A}

\begin{figure}[ht]
 \centering
 \includegraphics[scale=0.23]{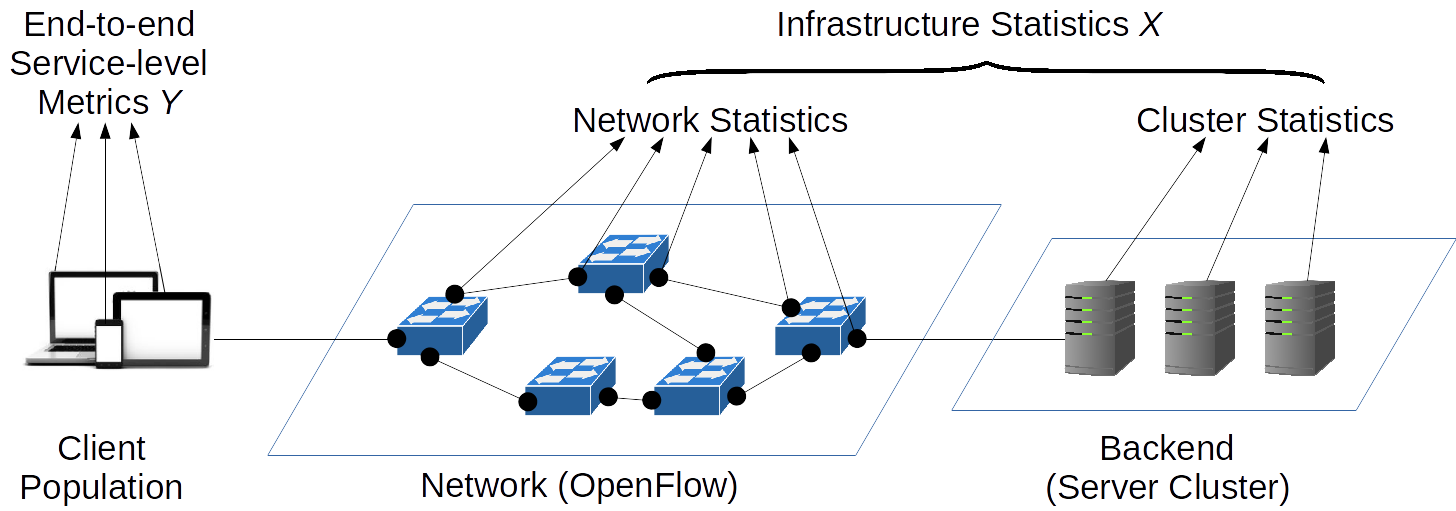}
 \caption{The testbed at KTH, providing the infrastructure for experiments. In various scenarios we predict end-to-end service-level metrics from low-level infrastructure measurements \cite{stadler2017learning}.}
 \label{fig:testbed}
\end{figure}

Figure \ref{fig:testbed} outlines our laboratory testbed at KTH. It includes a server cluster, an emulated OpenFlow network, and a set of clients. The server cluster is deployed on a rack with ten high-performance machines interconnected by a Gigabit Ethernet. Nine machines are Dell PowerEdge R715 2U servers, each with 64 GB RAM, two 12-core AMD Opteron processors, a 500 GB hard disk, and four 1 Gb network interfaces. The tenth machine is a Dell PowerEdge R630 2U with 256 GB RAM, two 12-core Intel Xeon E5-2680 processors, two 1.2 TB hard disks, and twelve 1 Gb network interfaces. All machines run Ubuntu Server 14.04 64 bits, and their clocks are synchronized through NTP \cite{NTP}.

\textit{The VoD service} uses VLC media player software \cite{VLC}, which provides single-representation streaming with varying frame rate. It is deployed on six PowerEdge R715 machines \textemdash one HTTP load balancer, three web server and transcoding machines, and two network file storage machines. The load balancer runs HAProxy version 1.4.24 \cite{haproxy}. Each web server and transcoding machine runs Apache version 2.4.7 \cite{apache} and ffmpeg version 0.8.16 \cite{ffmpeg}. The network file storage machines run GlusterFS version 3.5.2 \cite{gluster_fs} and are populated with the ten most-viewed YouTube videos in 2013, which have a length of between 33 seconds and 5 minutes. The VoD client is deployed in another PowerEdge R715 machine and runs VLC \cite{VLC} version 2.1.6 over HTTP.

\textit{The KV store service} uses the Voldemort software \cite{voldemort}. It executes on the same machines as the VoD service. Six of them act as KV store nodes in a peer-to-peer fashion, running Voldemort version 1.10.22 \cite{voldemort}.  
The OpenFlow network includes 14 switches, which interconnect the server cluster with clients and load generators. The load generators emulate client populations. 

A more detailed description of the testbed setup is given in \cite{stadler2017learning}. 
 
\subsection{Collected data and traces}
\label{subsec:collected_data}
We describe the metrics we collect on the testbed, namely, the input feature sets $\emph{\textbf{X}}_{cluster}$ and $\emph{\textbf{X}}_{port}$ \textemdash the union of which we refer to as $\emph{\textbf{X}}$ \textemdash as well as the specific service-level metrics $\emph{\textbf{Y}}_{VoD}$ and $\emph{\textbf{X}}_{KV}$.

\textit{The $\emph{\textbf{X}}_{cluster}$ feature set} is extracted from the kernel of the Linux operating system that runs on the servers executing the applications. To access the kernel data structures, we use System Activity Report (SAR), a popular open-source Linux library \cite{linux_sar}. SAR in turn uses procfs \cite{procfs} and computes various system statistics over a configurable interval. Examples of such statistics are CPU core utilization, memory utilization, and disk I/O. $\emph{\textbf{X}}_{cluster}$ includes only numeric features from SAR, about 1 700 statistics per server.

\textit{The $\emph{\textbf{X}}_{port}$ feature set} is extracted from the OpenFlow switches at per-port granularity. It includes statistics from all switches in the network, namely 1) Total number of Bytes Transmitted per port, 2) Total number of Bytes Received per port, 3) Total number of Packets Transmitted per port, and 4) Total number of Packets Received per port.

\textit{The $\emph{\textbf{Y}}_{VoD}$ service-level metric} is measured on the client device. During an experiment, we capture the \textit{Display Frame Rate} (frames/sec), i.e., the number of displayed video frames per second. This metric is not directly measured, but computed from VLC events like the display of a video frame at the client's display unit. We have instrumented the VLC software to capture these events and log the metric every second. 

\textit{The $\emph{\textbf{Y}}_{KV}$ service-level metric} is measured on the client device. During an experiment, we capture \textit{Read Response Time} as the average read latency for obtaining responses over a set of operations performed per second. This metric is computed using a benchmark tool of Voldemort, which we modified for our purposes. The read operation follows the request\textendash reply paradigm, which allows for tracking the latency of individual operation. We instrumented the benchmark tool to log the metric every second.

\textit{Generating the traces:} During experiments, $\emph{\textbf{X}}$ and $\emph{\textbf{Y}}$ statistics are collected every second on the testbed. For each application running on the testbed, the data collection framework produces a trace in form of a time series ${(\emph{\textbf{X}}_t, \emph{\textbf{Y}}_t)}$. We interpret this time series as a set of samples $\{(\emph{\textbf{X}}_1, \emph{\textbf{Y}}_1), ..., (\emph{\textbf{X}}_m, \emph{\textbf{Y}}_m)\}$.



\subsection{Generating load on the testbed}
\label{subsec:Generating_load}

We have built two load generators, one for the VoD application and another for the KV application. The VoD load generator dynamically controls the number of active VoD sessions, spawning and terminating VLC clients. The KV load generator controls the rate of KV operations issued per second. Both generators produce load according to two distinct load patterns.

\subsubsection{\textit{Periodic-load pattern}} the load generator produces requests following a Poisson process whose arrival rate is modulated by a sinusoidal function with starting load level $P_{S}$, amplitude $P_{A}$, and period of 60 minutes;
\subsubsection{\textit{Flash-crowd load pattern}} the load generator produces requests following a Poisson process whose arrival rate is modulated by the flash-crowd model described in \cite{flashcrowd}. The arrival rate starts at load level $F_{S}$ and peaks at flash events, which are randomly generated at rate $F_{E}$ events/hour. At each flash event, the arrival rate increases within a minute to a peak load $F_{R}$. It stays at this level for one minute and then decreases to the initial load within four minutes.

Table \ref{tab:parameters} shows the configurations of the load generators during the experiments reported in Section \ref{subsec:testbed_and_services}. We used a single load generator for the VoD experiments (see \cite{yanggratoke2015predicting}) and three for the KV experiments (see \cite{stadler2017learning}). 


\begin{table}[ht]
\centering
\caption{Configuration parameters of VoD and KV load generators.}
\label{tab:parameters}
\begin{tabular}{|c|c|c|c|c|c|c|}
\hline
\multirow{2}{*}{Application} & Load & \multicolumn{2}{|c|}{Periodic-load} &  \multicolumn{3}{|c|}{Flash-crowd-load}  \\\hhline{~~-----}
 & Generator & $P_{S}$ & $P_{A}$ & $F_{S}$ & $F_{E}$ & $F_{R}$ \\\hline
 \multirow{1}{*}{VoD} & 1 & 70 & 50 & 10 & 10 & 120 
\\\hhline{=======}
\multirow{2}{*}{KV} & 1 & 1\,000 & 800 & 200 & 10 & 1\,800 \\\hhline{~------}
 & 2, 3 & 350 & 150 & 200 & 3 & 500 \\\hhline{-------}
\end{tabular}
\end{table} 

\subsection{The scenarios chosen for this paper}
\label{subsec:scenarios}

The prediction method proposed in this paper has been evaluated using data from four experiments. Two of them involve running the VoD service and two the KV service.

\begin{enumerate}
\item \textit{VoD periodic}: In this experiment, we run the VoD service and generate a periodic load pattern on the testbed. Load generator and client are directly connected to the server cluster, and the testbed does not include the network (see Figure \ref{fig:testbed}). Data is collected every second over a period of $50\,000$ seconds. The $\emph{\textbf{X}}$ feature set contains $4\,984$ features. After cleaning the dataset, $50\,000$ samples and $1\,296$ features remain for processing. More details about the experiment are given in \cite{yanggratoke2015predicting}, and the trace is available at \cite{CNSM2019-traces}.

\item \textit{VoD flash-crowd}: This experiment relies on the same setup as VoD periodic, except that the testbed is loaded using the flash-crowd pattern. We process the trace the same way as described above and the given references contain more information. In this data set, after cleaning the dataset, $50\,000$ samples and $1\,255$ features remain for processing. 

\item \textit{KV periodic}:
In this experiment, we run the KV service to generate a periodic load pattern on the testbed. Unlike the VoD periodic experiment, we connect load generator and clients to the server cluster via an OpenFlow network (see\ref{sec:testbed-A}). Measurements are collected every second over a period of $28\,962$ seconds. The $\emph{\textbf{X}}_{cluster}$ feature set contains $10\,374$ features and the $\emph{\textbf{X}}_{port}$ feature set contains $176$ features. After cleaning the data set, $1\,751$ features remain for the processing. More details about the experiment are given in \cite{stadler2017learning}, and the trace is available at \cite{CNSM2019-traces}.

\item \textit{KV flash-crowd}:
This experiment relies on the same setup as KV periodic, except that the testbed is loaded using the flash-crowd pattern. We process the trace the same way as described above and the given references contain more information. In this data set, after cleaning the dataset, $19\,444$ samples and $1\,723$ features remain for processing. 

\end{enumerate}

In the evaluations reported in Sections V and VI, we perform two preprocessing steps on the trace data before they are read by OSFS. First, we use $MinMaxScaler$ to scale the values of each feature vector to the range $[0,1]$. Second, we remove the features with a variance below 0.0001.

%% file: computin_stabl_feature_sets.tex
\section{Computing stable feature sets}
\label{sec:Computing_stable_feature_sets}

In Section \ref{sec:creating_ranked_feature_lists} we discussed the algorithms ARR, LS, and TB, which produce ranked feature lists after reading $m$ samples. In an online setting, where samples become available one-by-one at discrete times $t=1,2,...$, the value of $m$ can be interpreted as time. Let's assume we run a feature ranking algorithm at time $t_1$ and compute the set with the top $k$ features, $\emph{\textbf{F}}_{k, t_1}$. This set will generally be different from the set $\emph{\textbf{F}}_{k, t_2}$ produced by the same  algorithm at a later time $t_2$. 
Using the standard assumption from statistical learning that samples are drawn from static distributions \cite{vapnik98}, we can assume that the sets $\emph{\textbf{F}}_{k, t}$ converge to a set $\emph{\textbf{F*}}_{k}$ with growing time $t$. Our objective for the remainder of the section is to find a heuristic criterion to determine at which time the feature sets $\emph{\textbf{F}}_{k, t}, t=1,2,3,...$ have sufficiently converged or, as we also say, have become 
``stable".

We first introduce a metric $sim()$ that captures the similarity between two sets A and B, which both contain $k>0$ elements:
\begin{equation}
    sim(A,B):=|A\cap B|/k
\end{equation}
The value of $sim$ is between 0 and 1. $0$ means no similarity between A and B, $1$ means maximum similarity, i.e. A and B are identical.

We study the evolution of $sim(\emph{\textbf{F}}_{k, t}, \emph{\textbf{F*}}_{k})$ over time using the data traces from our testbed. In this case, $\emph{\textbf{F*}}_{k}$  denotes a feature set that has been computed using an entire trace of samples (i.e. 20 000 - 50 000 samples, depending on the trace).

\begin{figure}[ht]
\centering
\subfigure[$sim(\emph{\textbf{F}}_{k,t},\emph{\textbf{F*}}_{k})$ vs $t$ in ARR]{
\begin{minipage}[t]{0.48\linewidth}
\centering
\includegraphics[width=1.7in]{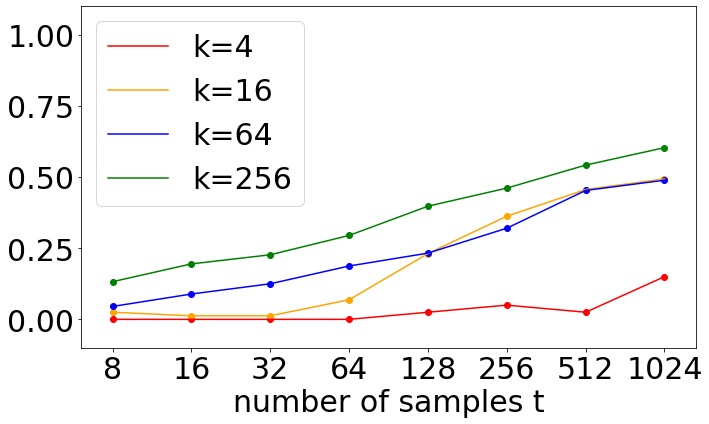}
\end{minipage}%
}%
\subfigure[$sim(\emph{\textbf{F}}_{k,t},\emph{\textbf{F*}}_{k})$ vs $t$ in LS]{
\begin{minipage}[t]{0.48\linewidth}
\centering
\includegraphics[width=1.7in]{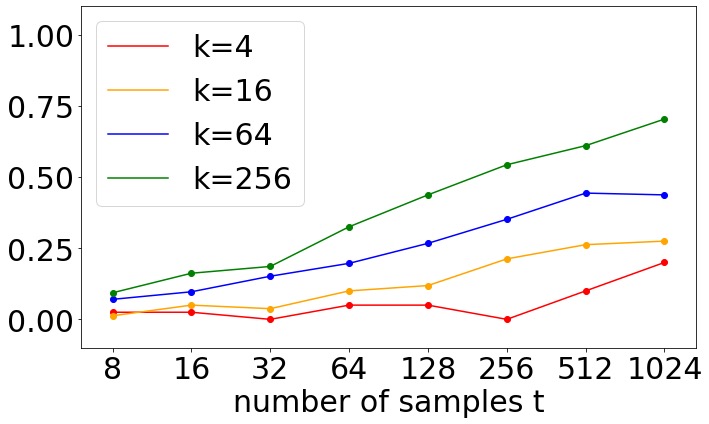}
\end{minipage}%
}%

\subfigure[$sim(\emph{\textbf{F}}_{k,t/2},\emph{\textbf{F}}_{k,t})$ vs $t$ in ARR]{
\begin{minipage}[t]{0.48\linewidth}
\centering
\includegraphics[width=1.7in]{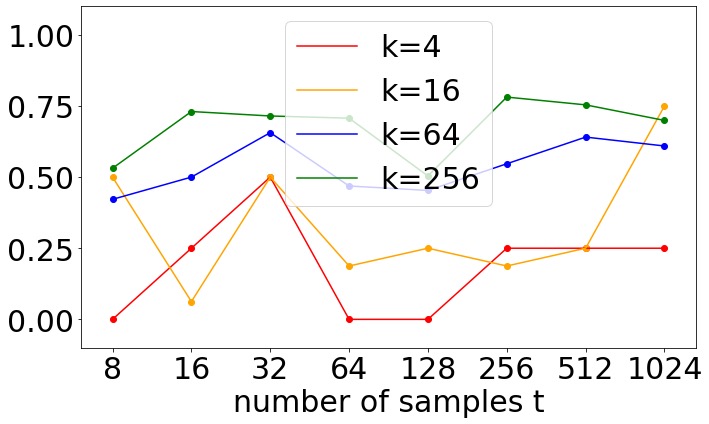}
\end{minipage}%
}%
\subfigure[$sim(\emph{\textbf{F}}_{k,t/2},\emph{\textbf{F}}_{k,t})$ vs $t$ in LS]{
\begin{minipage}[t]{0.48\linewidth}
\centering
\includegraphics[width=1.7in]{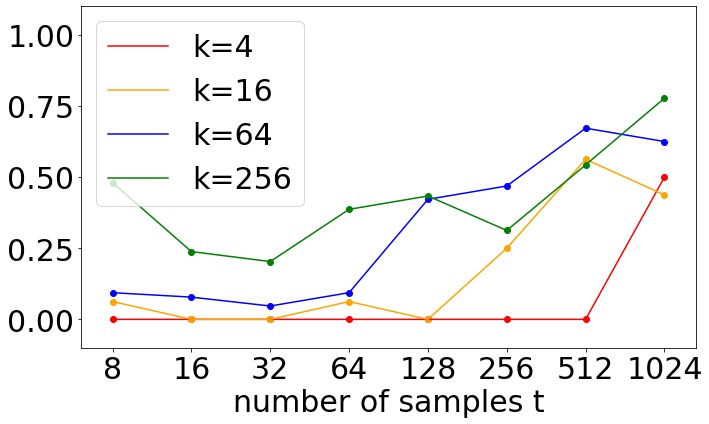}
\end{minipage}%
}%

\subfigure[NMAE vs $t$ in ARR]{
\begin{minipage}[t]{0.48\linewidth}
\centering
\includegraphics[width=1.7in]{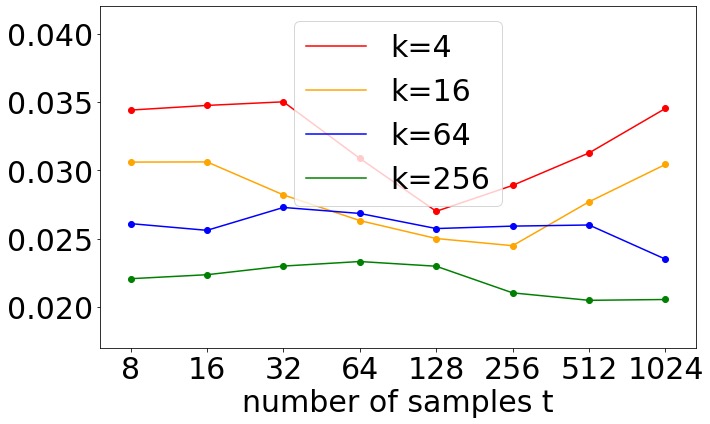}
\end{minipage}%
}%
\subfigure[NMAE vs $t$ in LS]{
\begin{minipage}[t]{0.48\linewidth}
\centering
\includegraphics[width=1.7in]{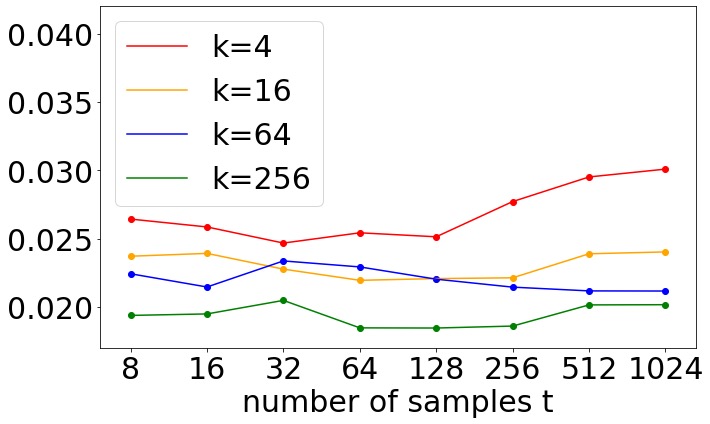}
\end{minipage}%
}%

\centering
\caption{The similarity of consecutive feature sets over time using ARR and LS with samples from the KV flash-crowd data set and start point $t=1$. The bottom row shows the prediction accuracy of a random-forest regressor using the feature set $\emph{\textbf{F}}_{k,t}$ for training.}
\label{pic:stability_s}
\end{figure}

Figure \ref{pic:stability_s} gives in the top row the value of $sim(\emph{\textbf{F}}_{k,t},\emph{\textbf{F*}}_{k})$ in function of time, computed by the ranking algorithms ARR and LS on the KV flash-crowd trace. The curves show the values for $k=4, 16, 64, 256$. Each point on the curves is the mean of 10 values from different start times, whereby one start time is t=1, and the other nine are chosen uniformly at random between t=2 and t=10 000. For each start time, $\emph{\textbf{F}}_{k,t}$ is computed and compared with $\emph{\textbf{F*}}_{k}$. (This applies to the top and bottom row. The curves in the middle row are based on a single start point with $t=1$ to better illustrate values encountered in an online scenario.)

As expected, the similarity values tend to increase over time, which means that the sets $\emph{\textbf{F}}_{k,t}$ and $\emph{\textbf{F*}}_{k}$ share more and more features as time progresses. We observe further that the similarity increases with increasing $k$.
Since in an online setting $\emph{\textbf{F*}}_{k}$ is not available for a very long time, we study the evolution of $sim(\emph{\textbf{F}}_{k,t},\emph{\textbf{F}}_{k, 2t})$ instead. The results are shown in the second row of Figure \ref{pic:stability_s}. We observe the same qualitative behavior of the curves as in the top row: the prediction error tends to decrease with increasing values of $k$ and $t$. 

The bottom row of the figure gives the prediction error of a random-forest regressor that has been trained using the feature set $\emph{\textbf{F}}_{k,t}$; it uses 100 trees. 
The prediction error is expressed as Normalized Mean Absolute Error (NMAE), which is computed as follows:
\begin{equation}
  NMAE=\frac{1}{\overline{y}}(\frac{1}{q}\sum_{i=1}^{q}\left | y_{i}-\widehat{y_{i}} \right |)  
\end{equation} 
$\overline{y}$ is the mean value of targets in test set,  q is the number of samples and $\widehat{y_{i}}$ is the predicted value.

From the evaluations presented in Figure 2 and additional ones we performed on other traces collected from our testbed, we decide on the following heuristic notion of a \emph{stable feature set}: 
\begin{center}
$\emph{\textbf{F}}_{k,t}$ is stable, if  $sim(\emph{\textbf{F}}_{k,t},\emph{\textbf{F}}_{k,2t}) > 0.5$
\end{center}
which means that $\emph{\textbf{F}}_{k,t}$ must share at least half of the features with $\emph{\textbf{F}}_{k,2t}$, which is computed with double the number of samples.

\begin{figure}[ht]
 \centering
 \includegraphics[scale=0.23]{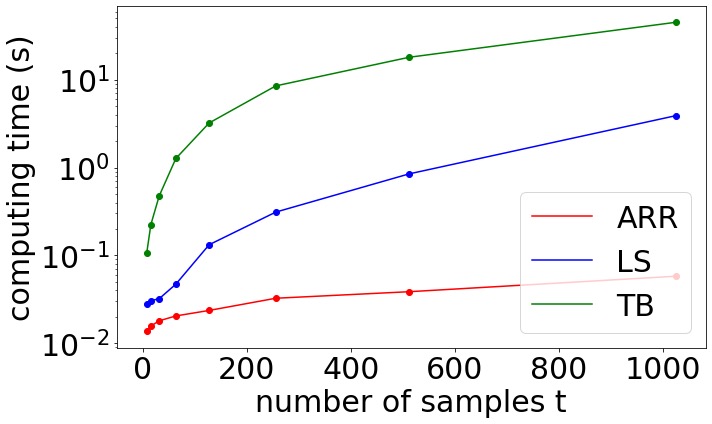}
 \caption{Time for computing $\emph{\textbf{F}}_{k,t}$ for all possible $k$ on a server using ARR, LS or TB. Data trace is  KV flash-crowd.}
 \label{fig:ct}
\end{figure}

Finally, some numbers on the computing time required to obtain the features sets. Figure \ref{fig:ct} gives the time to execute the feature ranking algorithms ARR, LS, and TB on a compute server of our testbed. $t$ is the number of samples the algorithms take as input. The time corresponds to computing the feature sets $\emph{\textbf{F}}_{k,t}$ for all possible $k=1...n$. To give a specific example, the execution time for processing 256 samples is 32 ms for ARR, 311 ms for LS, and 8532 ms for TB. The measurements suggest that computing $\emph{\textbf{F}}_{k,t}$ with ARR or LS in real time is feasible, also on smaller machines, e.g. on edge nodes.

%% file: online_feature_selection_with_low_overhead.tex
\section{Online feature selection with low overhead}
\label{sec:online_feature_selection_with_low_overhead}

In this section, we introduce the Online Stable Feature Set algorithm (OSFS), which reads a stream of samples and returns the number of features $k$, the number of samples $t_{k}$ needed to determine $k$, and the feature set $\emph{\textbf{F}}_{k}$. OSFS is instantiated with a feature ranking algorithm of choice. We evaluate the effectiveness of OSFS for several feature ranking algorithms using data traces from our testbed.

Recall that the purpose of OSFS is to select a subset of available data sources (i.e., features), in order to reduce the monitoring costs and the training overhead for learning. To keep costs and overhead low, we want $k$ and $t_k$ to be small while still allowing for effective learning and prediction.

The standard use case for OSFS plays out as follows. We start monitoring the values of $n$ features at time $t=1$ and collect a sequence of samples with index $t=1,2,... $ We use the collected samples to find small values for $k$ $(k<<n)$ and $t_{k}$ so that the first $t_{k}$ samples allow us to compute a stable feature set $\emph{\textbf{F}}_{t_k}$. At time  $t_{k}+1$, we reduce the number of data sources from $n$ to $k$ and continue motoring only sources from the set $\emph{\textbf{F}}_{k}$. Once we have collected $l$ samples, we take them as input to train the model for a learning task. The model is then applied for prediction using the samples collected after $t$=$l$. (In this work, we do not automatically determine the value for $l$ but set it to 1024 based on experience.)

\begin{algorithm} [ht]
\label{OSFS}
  \caption{Online Stable Feature Set (OSFS)}  
  \KwIn{Sample sequence $\emph{\textbf{X}}_{1}$, $\emph{\textbf{X}}_{2}$, $\emph{\textbf{X}}_{3}$,...;
  feature ranking algorithm $Rank$}  
  \KwOut{Feature subset $\emph{\textbf{F}}_k$, $k$, $t_{k}$\\
  $subset(k,t,Rank)$ returns top $k$ features computed with $Rank$ using samples with index $1,...,t$. }
 $\eta=0.5$ (threshold for stable feature set)\;
 $read=false$\;
 \For{$k$ in [4,16,64,256]}{
    \If{$not \ read$}{
    read and store $\emph{\textbf{X}}_{t}, t=1,...,16$
    }
  $\emph{\textbf{F}}_{k1}=subset(k,8,Rank)$\;
  $\emph{\textbf{F}}_{k2}=subset(k,16,Rank)$\;
  $sim_{k12}=sim(\emph{\textbf{F}}_{k1},\emph{\textbf{F}}_{k2})$;

  \For{$t= 17, ..., 1024$}
  {
    \If{$not \ read$}
    {
        read and store $\emph{\textbf{X}}_{t}$
   }
       \If{$t$ in [32,64,128,256,512,1024]}
    {

  $\emph{\textbf{F}}_{kt}=subset(k,t,Rank)$\;
    $sim_{kt}=sim(\emph{\textbf{F}}_{k2},\emph{\textbf{F}}_{kt})$\;
    \If{$sim_{kt}<sim_{k12}\ and\ sim_{k12}>\eta$}
    {
        return $\emph{\textbf{F}}_{k1},k,t/4$\;
    }
    \ElseIf{$sim_{kt}>\eta\ and\ t==1024$}{
    return $\emph{\textbf{F}}_{k2},k,t/2$\;
    }
    \Else{
    $\emph{\textbf{F}}_{k1}=\emph{\textbf{F}}_{k2}$\;
    $\emph{\textbf{F}}_{k2}=\emph{\textbf{F}}_{kt}$\;
    $sim_{k12}=sim_{kt}$;
    }
    }
    }
    $read=true$
       }
  return $\emph{\textbf{F}}_{k2},256,1024$  
\end{algorithm}

Algorithm \ref{OSFS} shows the pseudo-code of OSFS. The algorithm takes as input the sequence of arriving samples $\emph{\textbf{X}}_{t}$ and is initialized with a feature ranking algorithm, for instance $ARR$, $LS$, or $TB$. The ranking algorithm is used in the function $subset()$, which takes as input $k$ and the first $t$ samples and returns a set with the top $k$ features as ranked by the algorithm.
OSFS has two main loops: an outer loop (lines 3-23) that iterates over a subspace of $k$ and an inner loop (lines 9-22) that iterates over a subspace of $t$. The index values of $k$ and $t$ increase exponentially to enable exploration of a large space with a small number of evaluations. The algorithm performs a grid search in the space of tuples $(k,t)$ with the termination condition A: $sim_{kt}<sim_{k12}\ and\ sim_{k12}>\eta$ (line 15) or B: $sim_{kt}>\eta\ and\ t==1024$ (line 17). ($\eta$ is the threshold value for a stable feature set.) In case the conditions A or B are never met, the algorithm terminates after the search on the grid sector $[4-256]*[8-1024]$ has been completed. The key termination condition A expresses the case where (1) the similarity of two consecutive feature sets is above the threshold $\eta$ and (2) the similarity declines when the subsequent feature set is considered. 

OSFS reads at least 32 samples in order to ensure statistical viability and terminates after at most 1024 samples. 
Since the outer loop is indexed by increasing $k$, OSFS favors a smaller $k$ at the possible expense of a larger $t$. This means that we prefer reducing the monitoring overhead over reducing the time to compute $\emph{\textbf{F}}_{k}$.

We evaluate OSFS for the feature ranking algorithms ARR, LS, and TB using the four data traces described in \cite{XWFSRS2020}.

Table \ref{policy_nmae} summarizes the main results. The columns $k$ and $t_k$ give the figures for the feature selection algorithms we consider in this paper.  The values for $k$ and $t_k$ show the means and standard deviations from 10 start times for OSFS, whereby one start time is t=1 and the other 9 are chosen uniformly at random between t=2 and t=10\ 000. 

The columns NMAE1 and NMAE2 indicate the effectiveness of the feature sets produced by OSFS. They measure the error of a random-forest predictor using the produced feature sets. The values for NMAE1 relate to the case where the predictor models are trained using the 1024 samples after the respective start times, and the error is computed using all samples of the trace following those used for model training. In contrast, the values for NMAE2 relate to the case where the predictor models are trained in an offline fashion; i.e. 70\% of the samples of the trace is used for training and the rest for evaluation. ``No FS" refers to the case where all features are considered for training. Therefore, the column NMAE2 shows the effectiveness of the feature sets produced by OSFS compared to a baseline that considers all features. The column NMAE1 gives insights about how OSFS performs in conjunction with different feature selection algorithms for a specific learning task in a realistic online setting.

\begin{table*}[ht]
    \centering
    \caption{Evaluation of OSFS for feature selection algorithms ARR, LS, and TB for a the task of service metrics prediction on four data sets. The prediction method is random forest. No FS refers to the base line where all features are used by the predictor. NMAE1 is prediction accuracy where the prediction model is built with the first 1024 samples. NMAE2 is prediction accuracy where the prediction model is built through offline analysis with knowledge of the complete data set.}
    \label{policy_nmae}
    \begin{tabular}{|c|c|c|c|c|c|}
    \hline
    Dataset & Method &$k$ &$t_k$& NMAE1&NMAE2\\
    \hline
    KV flash-crowd&ARR&$25.6\pm 25.6$&$251.2\pm 224.0$&$0.0494\pm0.0010$&$0.0280\pm 0.0056$\\
    \cline{2-6}
    ~&LS&$10\pm 6$&$64\pm 72.3$&$0.0691\pm 0.0592$&$0.0232\pm 0.0014$\\
    \cline{2-6}
    ~&TB&$97.6\pm 106.6$&$332.8\pm 153.6$&$0.0221\pm 0.0012$&$0.0191\pm 0.0008$\\
    \cline{2-6}
    ~&No FS&1723&~&~&0.0184\\
    \hline
    KV periodic&ARR&$10\pm 6$&$249.6\pm 194.4$&$0.0573\pm0.0114$&$0.0325\pm 0.0071$\\
    \cline{2-6}
    ~&LS&$12.4\pm 5.5$&$195.2\pm 177.3$&$0.0711\pm 0.0473$&$0.0268\pm 0.0016$\\
    \cline{2-6}
    ~&TB&$156.4\pm 122.0$&$409.6\pm 159.9$&$0.0336\pm 0.0052$&$0.0232\pm 0.0028$\\
    \cline{2-6}
    ~&No FS&1751&~&~&0.0214\\
    \hline
    VoD flash-crowd&ARR&$12.4\pm 5.5$&$131.2\pm 142.7$&$0.1772\pm 0.0245$&$0.1629\pm 0.0139$\\
    \cline{2-6}
    ~&LS&$17.2\pm 16.5$&$226.4\pm 198.2$&$0.1631\pm 0.0354$&$0.0969\pm 0.0405$\\
    \cline{2-6}
    ~&TB&$54.4\pm 72.5$&$142.4\pm 198.2$&$0.1454\pm 0.0179$&$0.0999\pm 0.0306$\\
    \cline{2-6}
    ~&No FS&1255&~&~&0.0771\\
    \hline
    VoD periodic&ARR&$10\pm 6$&$363.2\pm 196.5$&$0.2172\pm0.0473$&$0.2085\pm 0.0203$\\
    \cline{2-6}
    ~&LS&$25.6\pm 25.6$&$108\pm 85.5$&$0.2145\pm 0.0527$&$0.1162\pm 0.0295$\\
    \cline{2-6}
    ~&TB&$47.2\pm 74.6$&$216\pm 242.2$&$0.2214\pm 0.0430$&$0.1473\pm 0.0429$\\
    \cline{2-6}
    ~&No FS&1296&~&~&0.119\\
    \hline 
    \end{tabular}
\end{table*}

The general conclusions we draw from Table \ref{policy_nmae} are as follows. Regarding the number of features $k$ and samples $t_k$:
\begin{itemize}
    \item OSFS instantiated with ARR, LS, or TB achieves a massive reduction of features  of 1-2 orders of magnitude. Most prominent is ARR, which produces feature sets with 10-20 features, down from 1255-1723 features.
    \item The number of samples needed to compute the feature sets averages around 100-400. For our tesbed, where metrics are monitored once per second, this means that the computed feature sets are available after 2-6 minutes of monitoring.
\end{itemize}

Regarding the accuracy of a specific predictor using the computed feature set:
\begin{itemize}
    \item Online OSFS with ARR, LS, or TB incurs a 50\%-100\% larger error than an offline-trained predictor with access to the entire feature set and the entire data trace. This is primarily due to the online computation of the feature sets. Further, comparing the columns NMAE1 and NMAE2 provides the cost of online prediction vs. offline prediction.
    \item Consistent with our earlier results (e.g., \cite{im2015_realm, stadler2017learning}), we find that the type of service and the load pattern significantly affect the prediction error.
\end{itemize}

When focusing on unsupervised feature selection, the comparison of ARR versus LS shows that ARR and LS exhibit similar performance regarding online feature selection and prediction (NMAE1). Since ARR has a significantly lower computing overhead (Table \ref{policy_nmae}, Figure \ref{fig:ct}), we conclude that ARR is the preferred algorithm in a resource-constrained environment (e.g. edge node or sensor node).

When feature selection is performed for a single predictor and target information is available, then TB should be considered for online feature selection. TB provides better prediction accuracy than ARR or LS, although at much higher computational costs (Table \ref{policy_nmae}, Figure \ref{fig:ct}) and sometimes larger $k$. 

From analysis and experimentation we know that increasing $t_k$, and to a lesser extent $k$, will improve the prediction accuracy, and OSFS can be extended for that purpose in a straightforward way. The particular technological use case determines the range of acceptable errors. For the services on our testbed, we believe we achieved monitoring and rapid training with low overhead at the cost of acceptable prediction errors.

%% file: related_work.tex
\section{Related work}
\label{sec:related_work}

Feature selection has been studied as an effective data prepossessing strategy in both the machine learning and data mining fields for several decades. Many survey papers cover and compare feature selection methods, e.g. \cite{guyon2003introduction}\cite{chandrashekar2014survey}\cite{li2017feature}\cite{solorio2020review}. For instance, \cite{li2017feature} provides a useful categorization of feature selection methods along several dimensions, such as supervised and unsupervised methods; wrapper, filter, and embedded methods; and static or streaming methods.

Most feature selection methods described in the literature assume a static feature space with a fixed number of instances (or samples), which makes them suitable for offline processing where all data fits into memory. In recent years however, online feature selection methods have received increasing attention. In this case, the input to the algorithm is either a stream of feature vectors (e.g. \cite{zhou2017online}\cite{alnuaimi2019streaming}\cite{wu2012online}\cite{perkins2003online}) or a stream of samples (e.g. \cite{wang2013online}\cite{shao2016online}), where data items are made available to the algorithm in a sequential fashion. Interestingly, most published online feature selection methods fall into the first category and a process a stream of feature vectors. Such methods are not suitable for the problem we study in this paper, since in the context of online monitoring the data becomes available as time progresses and, therefore, our interest is in methods that process streams of samples. Examples of such work are \cite{wang2013online} and \cite{shao2016online}. The authors of \cite{wang2013online} presents a supervised feature selection method for a linear classifier based on sparse online learning. More closely related to our problem domain is \cite{shao2016online}, which presents an unsupervised online feature selection method based on Non-negative Matrix Factorization (NMF) clustering. The method uses 
a fixed-size data structure for feature selection. Note though that both of these works do not address the questions of how many features should be selected (the value for $k$) or when to stop the algorithm (the value for $t_k$), both of which are central to our investigation.









%% file: conclusions_and_future_work.tex
\section{Conclusions and future work}
\label{sec:conclusions}

In this paper, we  introduced  an  online  algorithm,  OSFS,  that selects  a  small feature set from a large number of available data sources,  which  allows for  rapid,  low-overhead learning and prediction with acceptable errors. OSFS is instantiated with a feature ranking algorithm and performs a grid search that terminates when a stable feature set is identified and succeeding feature sets exhibit reduced similarity. 

Regarding future work, there are many ways our method can be improved and extended. In addition to the three ranking algorithms 
considered in this work, many other candidates that can be evaluated, together with OSFS, with respect to their ability to produce small values for $k$ and $t_k$ and predictors with acceptable errors. Of particular interest are online ranking algorithms that require only a single processing pass over the samples and whose memory requirement does not increase with $t$. Also, the tradeoffs between the three metrics $t_k$, $t_k$, and prediction error warrent further study. Finally, a formal analysis of the performance of OSFS with a specific ranking algorithm will be difficult due to dependencies on the particular data trace, but it should be attempted in order to fundamentally understand the conditions for efficient and effective online learning in networked systems.

%% file: acknowledgments.tex
\section{Acknowledgements}
\label{sec:ack}
The authors are grateful to Andreas Johnsson, Hannes Larsson, and Jalil Taghia with Ericsson Research for fruitful discussion around this work, as well as to Kim Hammar and Rodolfo Villa\c{c}a for comments on an earlier version of this paper. This research has been partially supported by the Swedish Governmental Agency for Innovation Systems, VINNOVA, through project AutoDC. 